\documentclass[letterpaper, conference]{IEEEtran}
\usepackage{amsmath,amssymb,amsfonts}
\usepackage{algorithmic}
\usepackage{graphicx}
\usepackage{textcomp}
\usepackage{xcolor}
\usepackage[numbers]{natbib}
\usepackage{multirow}
\usepackage{float}
\setcitestyle{square}

\IEEEoverridecommandlockouts

\usepackage{xcolor}

\def\BibTeX{{\rm B\kern-.05em{\sc i\kern-.025em b}\kern-.08em
    T\kern-.1667em\lower.7ex\hbox{E}\kern-.125emX}}
\begin{document}

\title{Evolving Spiking Neural Networks to Mimic PID Control for Autonomous Blimps\\
\thanks{This material is based upon work supported by the Air Force Office of Scientific Research under award number FA8655-20-1-7044.}
}

\author{\IEEEauthorblockN{Tim Burgers}
\IEEEauthorblockA{\textit{Delft University of Technology} \\
Delft, The Netherlands \\
}
\and
\IEEEauthorblockN{Stein Stroobants}
\IEEEauthorblockA{\textit{Delft University of Technology} \\
Delft, The Netherlands \\
s.stroobants@tudelft.nl
}
\and
\IEEEauthorblockN{Guido C.H.E. de Croon}
\IEEEauthorblockA{\textit{Delft University of Technology} \\
Delft, The Netherlands\\
g.c.h.e.decroon@tudelft.nl
}}

\maketitle

\begin{abstract}
In recent years, Artificial Neural Networks (ANN) have become a standard in robotic control. However, a significant drawback of large-scale ANNs is their increased power consumption. This becomes a critical concern when designing autonomous aerial vehicles, given the stringent constraints on power and weight. Especially in the case of blimps, known for their extended endurance, power-efficient control methods are essential. Spiking neural networks (SNN) can provide a solution, facilitating energy-efficient and asynchronous event-driven processing. 
In this paper, we have evolved SNNs for accurate altitude control of a non-neutrally buoyant indoor blimp, relying solely on onboard sensing and processing power. The blimp's altitude tracking performance significantly improved compared to prior research, showing reduced oscillations and a minimal steady-state error. The parameters of the SNNs were optimized via an evolutionary algorithm, using a Proportional-Derivative-Integral (PID) controller as the target signal. We developed two complementary SNN controllers while examining various hidden layer structures. The first controller responds swiftly to control errors, mitigating overshooting and oscillations, while the second minimizes steady-state errors due to non-neutral buoyancy-induced drift. Despite the blimp's drivetrain limitations, our SNN controllers ensured stable altitude control, employing only 160 spiking neurons. 
\end{abstract}


\section{Introduction}
Throughout history, humans have been fascinated by how animals gracefully and precisely navigate complex environments. This fascination has inspired efforts to understand the brain's computational processes behind such behavior, leading to the development of Artificial Neural Networks (ANN). These ANNs represent the neural processes using simplified mathematical models. Their ability to approximate complex non-linear functions makes them highly effective in controlling complex systems, such as quadrotors~\cite{Zhang2022AController, Heryanto2017}.
However, as the size of ANNs grow, both response latency and computational demands increase. The latter poses particular challenges for robotic applications with restricted onboard energy capacity, such as flying robots. A solution may be found in the information transmission methods employed by biological brains. 
ANNs rely on continuous-valued signals, whereas the biological brain employs sparse spatial-temporal "spike" signals—brief, rapid increases in neuron voltage—for data encoding and transmission.

\begin{figure}[t]
  \centering
  \includegraphics[width = 1.0\linewidth]{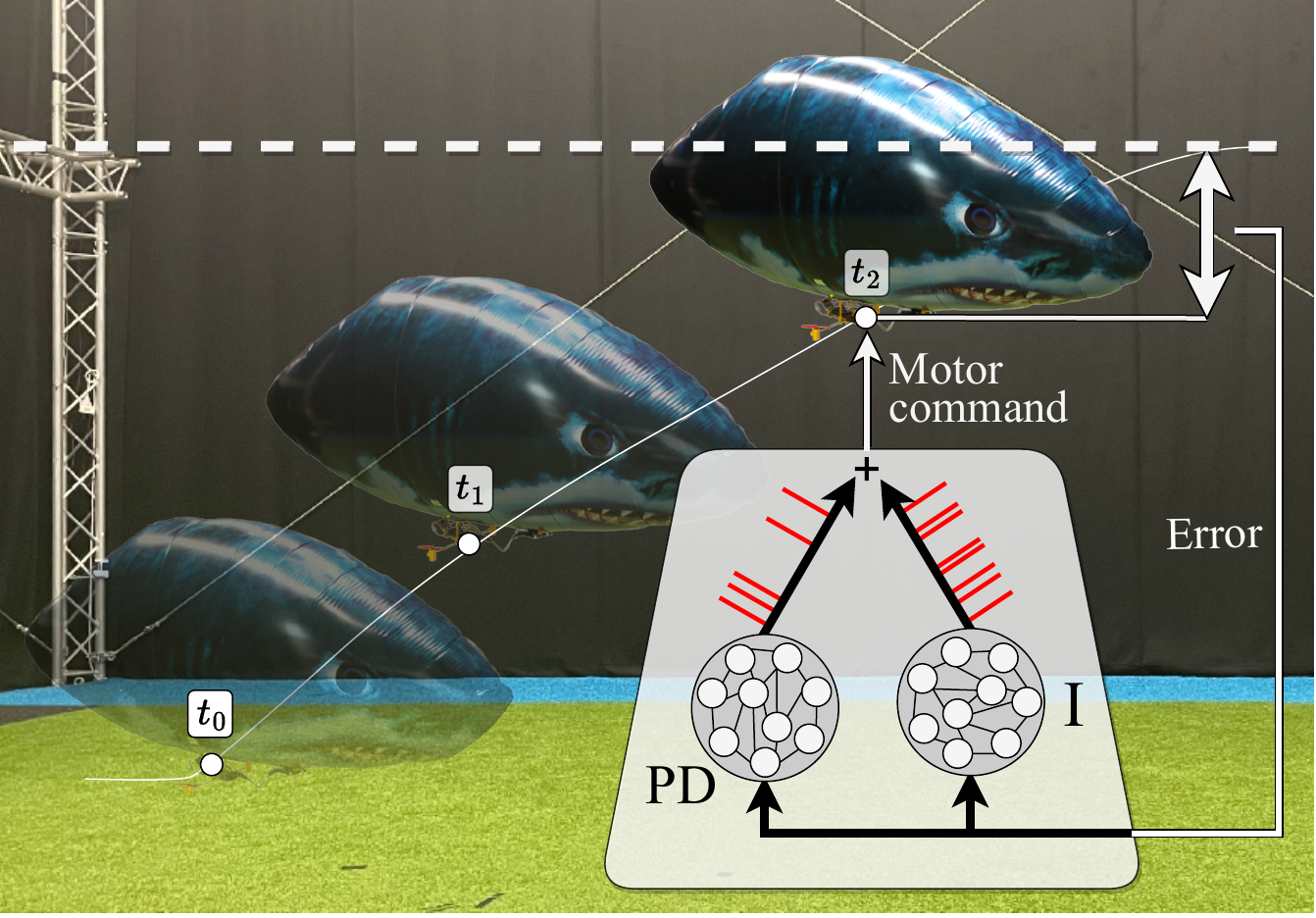}
  \caption{The proposed SNN altitude controller for the blimp, where [$t_0$,$t_1$,$t_2$] indicates time instances of the blimp's altitude while approaching a setpoint, marked by the dashed line.}
  \label{fig:front}
\end{figure}

There are neural networks that use this spike-based approach to transmitting information, called a Spiking Neural Network (SNN)~\cite{Maass1997NetworksModels}. These more biologically plausible neural networks show potential for energy-efficient and low-latency controllers~\cite{Bing2019ANetworks}. This fact was demonstrated in a study by Vitale et al.~\cite{Vitale2021Event-drivenChip}, where an SNN controller in a fully neuromorphic control loop outperformed a conventional control loop on power consumption and control latency when tracking the roll angle of a bench-fixed 1 DoF birotor. However, the application of SNN controllers in robotics is still in its early stages of development. One of the biggest challenges is the availability of suitable training algorithms. 
The SNN's temporal dynamics, sparse spiking activity, and non-differentiable spike signal make most existing ANN training algorithms unsuitable~\cite{Taherkhani2020ANetworks}. 
Recent research has enabled the use of error backpropagation for SNNs by means of surrogate-gradient algorithms~\cite{Neftci2019SurrogateNetworks}. Nevertheless, training SNNs using these gradient-based algorithms is still difficult due to the susceptibility to local minima, exploding/vanishing gradient, and sensitivity to initial conditions~\cite{Bing2019}.

Due to the increased training complexity of SNNs compared to non-spiking counterparts, practical applications of SNNs in the robotic control domain are still limited. Designing a fully neuromorphic SNN controller to emulate low-level controllers, such as the Proportional-Integral-Derivative (PID) control, still remains a complex task. Recent research showed SNNs performing differentiation and integration within the network, by manually configuring neuron connections and weights~\cite{Stagsted2020Event-basedStudy, Stagsted2020TowardsUAV, Stroobants2022DesignProcessors}.
In these studies, the controllers were implemented on Intel's Loihi neuromorphic processor, showcasing the promise of neuromorphic hardware by exhibiting very low latencies~\cite{Davies2018Loihi:Learning}. 
In Zaidel et al.~\cite{Zaidel2021NeuromorphicControl}, multiple populations with pre-determined network parameters were used to implement all three pathways of the PID controller to control a 6 Degrees of Freedom (DoF) robotic arm. The integral pathway was implemented using a fully recurrent population of neurons, while differentiation was achieved by using a slow and fast time constant for two populations. Although the integral controller succeeded in reducing the steady-state error, it was unable to completely eliminate it. In another study, spiking neurons were trained to replace the rate controller of a tiny quadrotor, the Crazyflie~\cite{Stroobants2023NeuromorphicAdaptation}. Integration in the SNN controller was achieved through discretized Input-Weighted Threshold Adaptation (IWTA), where the threshold depends on the previous layer's spiking activity. The training process had limitations because it relied partially on predetermined connections and grouped network parameters.

In this work, we investigate the different mechanisms, recurrency and IWTA, used in prior SNN PID research that enabled differentiation and/or integration.
For each mechanism, we evolve an SNN to control the altitude of a real-world indoor blimp. The blimp is an interesting test platform for the SNN controller, allowing validation of all components of the PID controller. The changing buoyancy over time requires a good integrator to be present. Moreover, due to the high delays and slow system dynamics of a blimp, a high proportional gain is necessary in the reference PID controller. This reduces the blimp's rise time, requiring a strong derivative in the controller's output to prevent overshoot and oscillations. In Gonzalez et al.~\cite{Gonzalez-Alvarez2022EvolvedBlimp}, an open-source indoor blimp was designed and used as a test vehicle for an evolved neuromorphic altitude controller, showing adequate tracking of the reference signal. However, even after including an additional non-spiking PD controller to the output of the SNN controller, there were still oscillations present of approximately $\pm$ 0.3m. Additionally, the SNN was only trained on a neutrally buoyant blimp. Slight changes in the buoyancy of the blimp would cause a steady state error, which the controller was unable to eliminate.

We build further on this research, presenting here the following contributions: 1) We developed a fully neuromorphic height controller for a blimp (visualized in Figure~\ref{fig:front}), using an evolved SNN of only 160 neurons that is able to minimize the overshoot and oscillations while also removing the steady-state error caused by the buoyancy of the blimp.
2) We analyze the individual and combined influence of recurrent connections and IWTA on the performance of the SNN controller 3) We made improvements to the hardware components of the open-source blimp, improving the onboard computational power and increasing the accuracy of the height measurements.


\section{Methodology}

The SNN controller consists of three layers of neurons, where all parameters are optimized using an evolutionary algorithm to mirror the output of a tuned PID controller. The Proportional-Derivative (PD) controller's rapid dynamics demand fast time constants, while the integral controller relies on slower dynamics and, thus, slow time constants. To facilitate the learning process, we split the controller into two separate parts based on the required time constants to model each component. After completing the training process, the evolved controllers are used to control the altitude of a helium-filled blimp. Detailed discussions on the SNN's structure, parameters, experiment setup, and the evolutionary training algorithm used in this study follow below.

\subsection{Spiking Neural Network Controller}
We use current-based Leaky-Integrate and Fire (LIF) neurons with a soft reset for the threshold ($\vartheta$). The discretized equations that describe the dynamics of the three states of the neuron (synaptic current $i(t)$, membrane potential $v(t)$ and spike train $s(t)$) are described as follows:
\begin{align}
    i_i(t)&=\tau_i^{\mathrm{syn}} i_i(t-1)+\sum W_{i j} s_j(t) \\
    v_i(t)&=\tau_i^{\mathrm{mem}} v_i(t-1)+i_i(t) - s_i(t-1) \vartheta_i \\
    s_i(t)&=H\left(v_i(t)-\vartheta_i\right)
\end{align}
where subscript $i$ and $j$ denote the post- and presynaptic neuron respectively. The discretized time constants, known as the decay parameters, of the synapses and the membrane potential are respectively referred to as $\tau^{syn}$ and $\tau^{mem}$. The spiking behavior of a neuron is modeled using the Heaviside step function $H$, which outputs a spike when the membrane potential exceeds the threshold.

\begin{figure}[h]
  \centering
  \includegraphics[width = 1.0\linewidth]{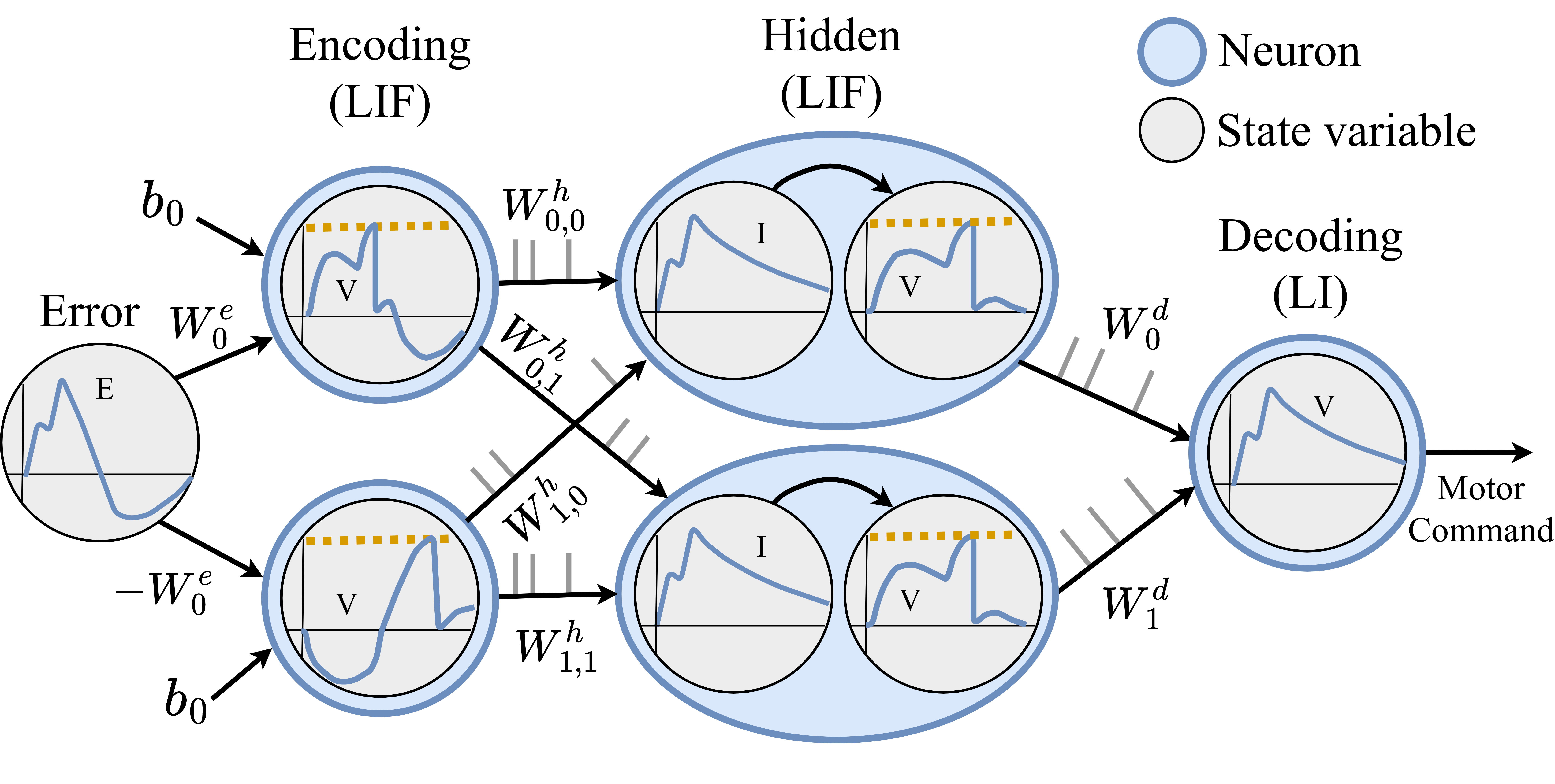}
  \caption{The basic structure of the SNN controller. The encoding layer has an additional bias ($b$) added to the input current.}
  \label{fig:snn_structure}
\end{figure}

The basic structure of a spiking neural network controller is schematically depicted in Figure~\ref{fig:snn_structure}. The input weights, indicated by $W^e$, $W^h$ and $W^d$, are linked to the encoding, hidden and decoding layers, respectively.  The encoding layer is responsible for translating the floating-point input into a sequence of spikes, and conversely, the decoding layer performs the reverse operation. 

\begin{figure}[h]
  \centering
  \includegraphics[width = 0.7\linewidth]{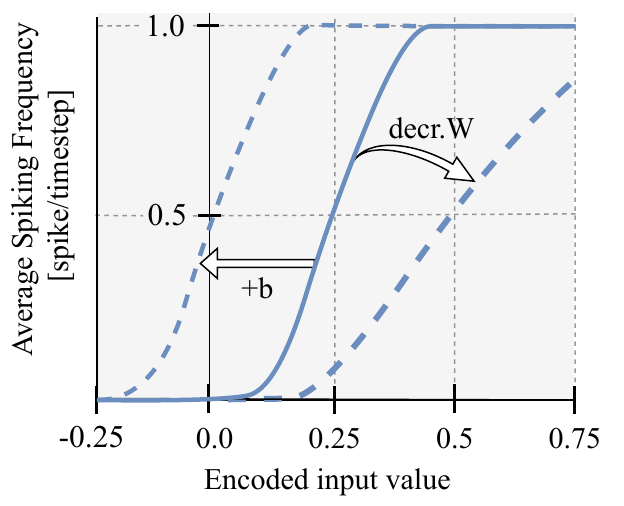}
  \caption{Influence of the input weight and bias on the spiking frequency of a LIF neuron.}
  \label{fig:encoding_response}
\end{figure}

The input to the network is the error of the controlled state. After applying the encoding weights and biases, the error signal is directly used as the synaptic current for the encoding neuron ($\tau^{syn}$ = 0). To facilitate the training of the encoding layer, we paired neurons with shared bias and flipped weight sign. This results in symmetric encoding, ensuring similar spiking patterns for both positive and negative errors. The effect that the input weight and bias have on the spiking behavior of a LIF neuron is presented in Figure~\ref{fig:encoding_response}. 
The encoding layer incorporates a bias to achieve spike activity for the encoding of error values close to zero, which would otherwise be impossible.

\begin{figure}[h]
  \centering
  \includegraphics[width = 0.9\linewidth]{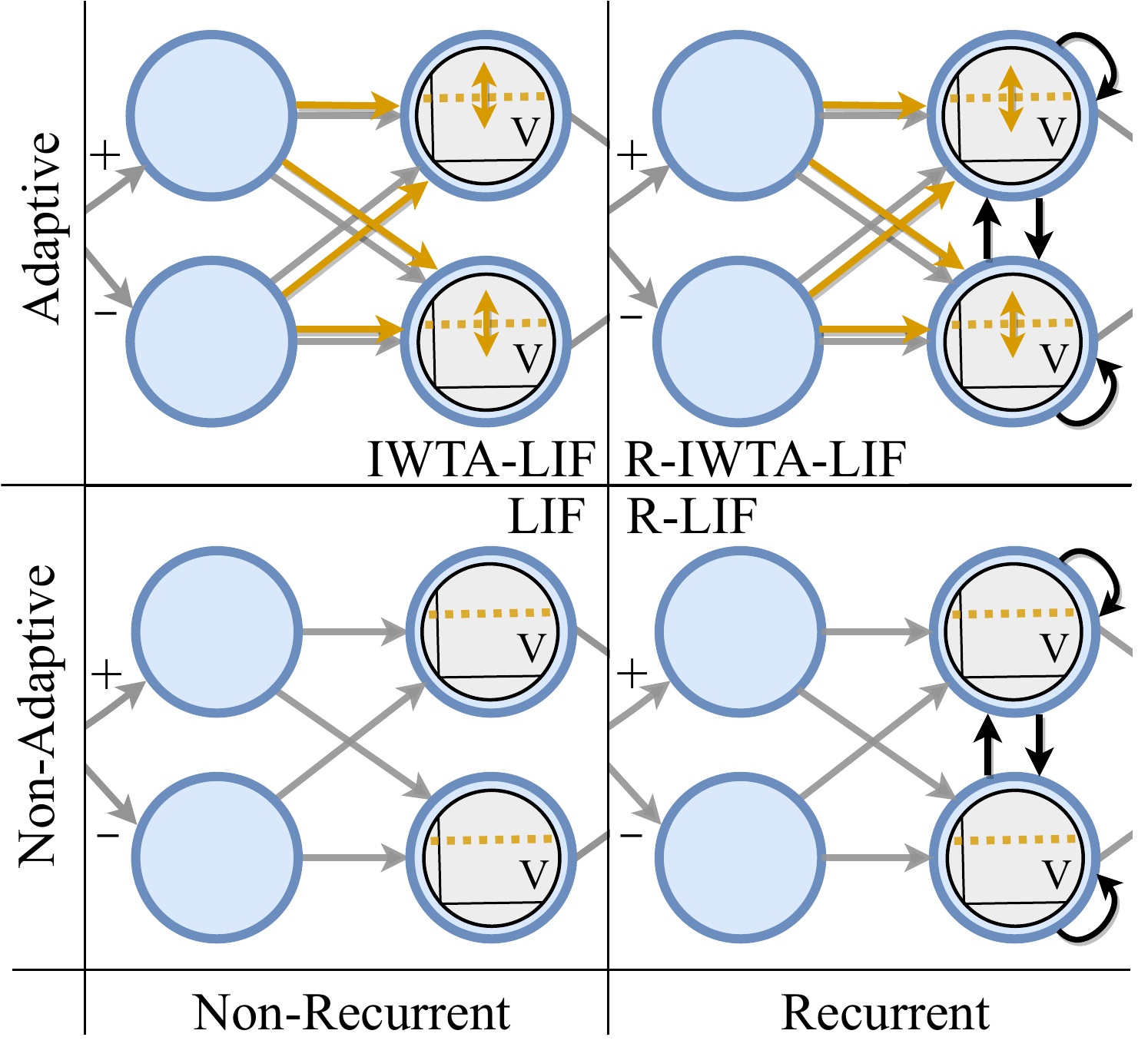}
  \caption{Visualization of the different hidden layer structures. The two left and right circles represent the encoding and the hidden neurons respectively.}
  \label{fig:hidden}
\end{figure}

We study the effect of using different types of structures for the hidden layer of the SNN controller in this work. The most basic type of hidden layer (LIF) is depicted in Figure~\ref{fig:snn_structure}. In addition to the basic structure, we evaluated the influence of recurrency~\cite{Zaidel2021NeuromorphicControl, Qiu2020a} and Input-Weighted Threshold Adaptation (IWTA)~\cite{Stroobants2023NeuromorphicAdaptation}, as both these network structures demonstrated their essential role in enabling integration within the SNN. We focused solely on threshold adaptation linked to the incoming spiking activity rather than the hidden neuron's activity itself. 
An overview of the different hidden layer structures is shown in Figure~\ref{fig:hidden}. In contrast to the original implementation in \cite{Stroobants2023NeuromorphicAdaptation}, the threshold for the IWTA-LIF neurons is modeled using a decay term ($\tau^{th}$), where the threshold converges back to a base value ($\vartheta$), after an increase/decrease ($W^{th}$) caused by an incoming spike. This method of implementing IWTA adds new dynamics to the threshold and increases the solution space.

The spiking neural network controller is decoded using a single leaky-integrator neuron, that calculates the exponential moving average of the spikes in the hidden layer.

\subsection{Real-World Experiment}

To validate the performance of the SNN controller on a real-world application, we implemented an SNN altitude controller for an open-source micro-blimp developed in~\cite{Gonzalez-Alvarez2022EvolvedBlimp}. The combination of the buoyancy-caused drift and slow system dynamics make the blimp a useful test vehicle for this research. 
\begin{figure}[b]
  \centering
  \includegraphics[width = 0.9\linewidth]{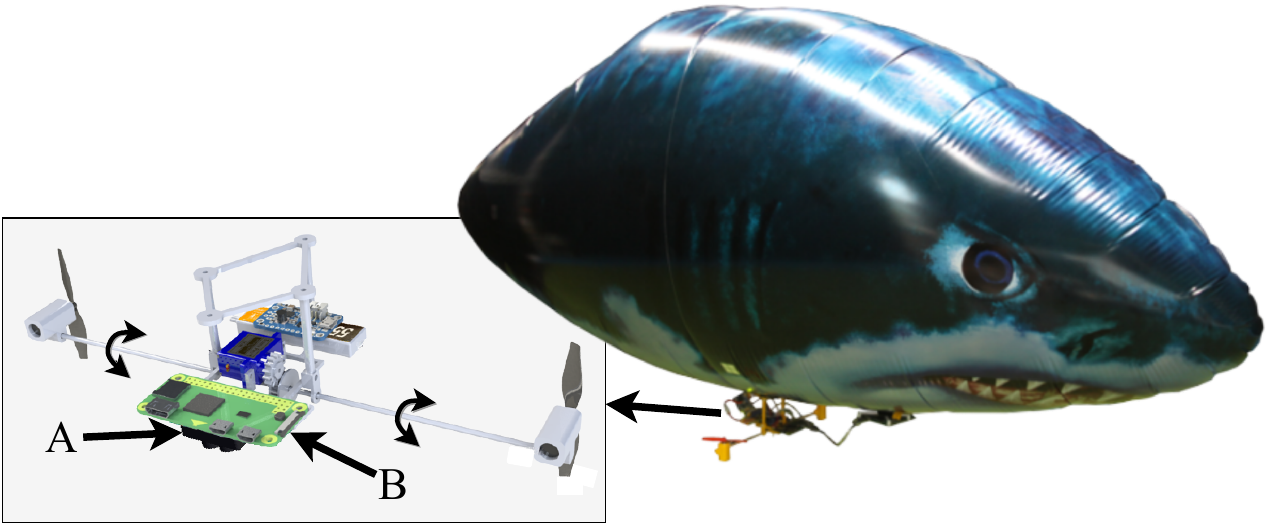}
  \caption{The open-source micro-blimp used for the real-world experiments~\cite{Gonzalez-Alvarez2022EvolvedBlimp}. Two adjustments made on the blimp are: A) TFmini S LiDAR-module B) Raspberry Pi Zero 2 W.}
  \label{fig:blimp}
\end{figure}
The input of the SNN height controller is the difference between a reference altitude and the onboard lidar measurement, which makes it a fully on-board closed-loop control system. The output of the SNN is the motor command, $u \in [-3.3,3.3]$, where $u<0$ indicates downward and $u>0$ upwards movement. 
The blimp's control system consists of two coreless direct current (DC) motors attached to a 180-degree rotating shaft, which enable the rotors to push the blimp both up- and downwards. A visual representation of the blimp and its hardware components is provided in Figure~\ref{fig:blimp}. Two improvements have been made to the blimp's hardware setup. Firstly, to improve the altitude tracking ability, we implemented a LiDAR sensor, the TFmini S LiDAR-module, which significantly increased the accuracy from $\pm$ 20cm to $\pm$ 1cm. Secondly, the Raspberry Pi Zero 2 W, running on Ubuntu 20.04, replaced its predecessor to increase the processing power and prevent software compatibility issues. The total weight of all hardware components attached to the blimp is 140g. In order to ensure smooth communication between all system components, we have used the Robot Operating System (ROS1) framework. The control loop runs at a rate of 10 Hz.

\subsection{Evolutionary Training Algorithm}
The training process for the SNN controller employs an evolutionary algorithm, consisting of two recurring steps: population creation and evaluation, with each cycle representing one generation of the evolution. 
In the first step, a set number of individuals forms the population, each representing a unique controller with slightly varying parameters. The second step ranks these individuals based on performance via a cost function, and this information guides the creation of the new population. Further details for each step are discussed below. 

Every training loop was executed on the DelftBlue supercomputer, running on 20 cores for 50,000 generations with 50 individuals~\cite{DHPC2022_new}. To prevent overfitting, we randomly sampled one of the 100 training datasets to evaluate each generation. Additionally, both the size and the frequency of the step inputs used in each dataset were varied to enhance the diversity of the training data. We conducted a total of 30 training loops for each combination of controller type (2) and hidden layer structure (4), resulting in a total of 240 training sessions. Each controller is limited to 80 neurons to showcase the potential of small-scale networks for neuromorphic control. 

\subsubsection{Population Creation}
For the creation of a new population, we have used a Covariance Matrix Adaptation Evolutionary Strategy (CMA-ES)~\cite{hansen2006cma}. CMA-ES is a distribution-based optimization algorithm that iteratively adjusts the mean and covariance matrix of a multi-variate Gaussian distribution, from which all network parameters of each individual are sampled. 

The CMA-ES is implemented using the \textit{Evotorch} Python library~\cite{Toklu2023EvoTorch:Python}. Every network parameter that needs to be evolved is initialized using  the mean and standard deviation of a Gaussian distribution. Strict-bounded parameters, such as the decay constants in the neuron model, are constrained by rejecting and resampling. The initial mean is set by sampling from a uniform distribution within parameter bounds, while the standard deviation is set to 1/10th of the parameter's range. An overview of all trained parameters including the bounds is provided in Table~\ref{tab:parameters_controller}, where $W^r$ represents the recurrent weights.
\begin{table}[h!]
\caption{Overview of all parameters and bounds used to evolve the SNN, with additional parameters in the hidden layer for $^\ast$recurrency and $^\dagger$IWTA}
\begin{tabular}{lllllc|l|l|l|}
\multicolumn{1}{l|}{}                      & \multicolumn{1}{l|}{Param.}   & \multicolumn{1}{l|}{Size} & \multicolumn{1}{l|}{Bounds}       &  & \multicolumn{1}{l|}{}   & Param.     & Size  & Bound       \\ \cline{1-4} \cline{6-9} 
\multicolumn{1}{l|}{\multirow{4}{*}{Enc.}} & \multicolumn{1}{l|}{$W^e$}    & \multicolumn{1}{l|}{N}    & \multicolumn{1}{l|}{{[}-2, 2{]}} &  & \multirow{7}{*}{\rotatebox{90}{Hidden}} & $W^h$      & 2N    & {[}-2, 2{]} \\ \cline{2-4} \cline{7-9} 
\multicolumn{1}{l|}{}                      & \multicolumn{1}{l|}{$b$}     & \multicolumn{1}{l|}{N}    & \multicolumn{1}{l|}{{[}-1, 1{]}} &  &                         & $\tau^{syn}$   & N     & {[}0, 1{]}  \\ \cline{2-4} \cline{7-9} 
\multicolumn{1}{l|}{}                      & \multicolumn{1}{l|}{$\tau^{mem}$} & \multicolumn{1}{l|}{N}    & \multicolumn{1}{l|}{{[}0,1{]}}   &  &                         & $\tau^{mem}$   & N     & {[}0, 1{]}  \\ \cline{2-4} \cline{7-9} 
\multicolumn{1}{l|}{}                      & \multicolumn{1}{l|}{$\vartheta$ }    & \multicolumn{1}{l|}{N}    & \multicolumn{1}{l|}{{[}0,10{]}}  &  &                         & $\vartheta$      & N     & {[}0, 10{]} \\ \cline{1-4} \cline{7-9} 
                                           &                               &                           &                                  &  &                         & $^\ast W^{r}$ & NxN & {[}-1, 1{]} \\ \cline{1-4} \cline{7-9} 
\multicolumn{1}{l|}{\multirow{2}{*}{Dec.}} & \multicolumn{1}{l|}{$W^d$}    & \multicolumn{1}{l|}{N}    & \multicolumn{1}{l|}{{[}-1,1{]}}  &  &                         & $^\dagger \tau^{th}$ & N     & {[}0,1{]}   \\ \cline{2-4} \cline{7-9} 
\multicolumn{1}{l|}{}                      & \multicolumn{1}{l|}{$\tau^{mem}$} & \multicolumn{1}{l|}{N}    & \multicolumn{1}{l|}{{[}0,1{]}}   &  &                         & $^\dagger W^{th}$  & NxN & {[}-1, 1{]} \\ \cline{1-4} \cline{6-9} 
\end{tabular}
\label{tab:parameters_controller}
\end{table}

\subsubsection{Population Evaluation}
The performance of each individual SNN controller is determined by comparing the output of the SNN controller ($u$) to the output of a tuned PD or integral controller ($\hat{u}$). The SNN is tasked to learn the mapping between the input signal ($e$) to the output of the target controller ($\hat{u}$). The discretized equation that describes the PID response is provided below:
\begin{equation}
    \hat{u_k}= \underbrace{K_p e_k+ K_d\frac{e_k-e_{k-1}}{T}}_\text{PD controller} + \underbrace{K_i(\sum_{k=0}^{k}T e_k)}_\text{Integral controller}
\end{equation}
where $K_p$, $K_i$, and $K_d$ refer to the proportional, integral and derivative gains respectively and  $T$ is denoted by the sampling period. Both the input error signal and the target controller response are recorded in a dataset. 
 

To quantify the performance of the SNN compared to the target controller, the Mean Absolute Error (MAE) was used as the main term in the cost function. The MAE was used instead of the mean square error (MSE) to prevent over-penalization of the error that is created during the transient response to a step input because this led to more oscillations in the steady state. Additionally, the cost function was augmented with the Pearson Correlation Coefficient (PCC)~\cite{Benesty2009}, denoted by $\rho$, to incentivize that the sign of the output of the SNN and PD/I controllers is equal. The PCC measures the linear correlation between two signals, ranging from -1 to 1, where the latter indicates a fully linear relation. Since the fitness function is a minimization function, the PCC is included by adding $1-\rho$ to the MAE. This results in the following cost function that was used in the population evaluation step of the evolutionary training process:
\begin{equation}
    L(u,\hat{u}) = \text{MAE}(u,\hat{u}) + (1-\rho(u,\hat{u}))
\end{equation}

\subsection{Dataset Generation}
The methods used to gather the test and training data for each controller are discussed below.

\subsubsection{PD controller}
To successfully train the PD SNN controller, we need a broad spectrum of error signals. Therefore, we used a semi-randomly tuned PID controller on the neutrally buoyant real-world blimp. The error signal of these recordings is used as the SNN input data of the dataset. The target signal for the training algorithm is generated by passing the recorded error signal through a PD controller with the tuned gains for the blimp's altitude controller. 



\subsubsection{Integral controller}
The Integral SNN has to learn to integrate the error within the SNN itself. For this training process, the decay parameter of the decoding neuron is purposely bound to ensure a quick decay (eg. [0-0.3]) in order to prevent the algorithm from converging to a slow decay. A slow decoding decay parameter would imply that the integration only happens in the decoding neuron, instead of in the hidden layer.

If the buoyancy remains constant throughout the training process of the integral controller, the network might learn to add a bias to counteract the buoyancy. Therefore we must adjust the buoyancy during training to ensure that the network learns to integrate information over time. Instead of recording multiple datasets with varying buoyancy, we decided to train the Integral controller on a model where we could also change the "buoyancy" within a single dataset to facilitate the learning process.  

The blimp was modeled by the double integrator control problem and controlled using a PID controller. The integral gain was set to match that of the tuned real-world blimp, while the PD gains were adjusted to align the model's dynamics approximately with those of the tuned PID-Blimp system. In the double integrator system, the output of the controller, $u(t)$ is directly proportional to the second derivative of the state plus an additional bias: $\ddot{x}(t) = u(t) + b$ . The bias simulates the level of buoyancy of the blimp. Every time a step input is received, the bias is randomly sampled ($U(-4,4)$). Each dataset consists of 5 step inputs maintained for 50s.

\section{Results}
The first subsection displays the training outcome of both SNN controllers using a test dataset, followed by the assessment of their performance on an actual blimp. The results also contain a performance analysis of various neural mechanisms within the hidden layer. 

\subsection{Training of the SNN Controllers}


\subsubsection{PD SNN Controller}

\begin{figure}[H]
  \centering
  \includegraphics[width = 1.0\linewidth]{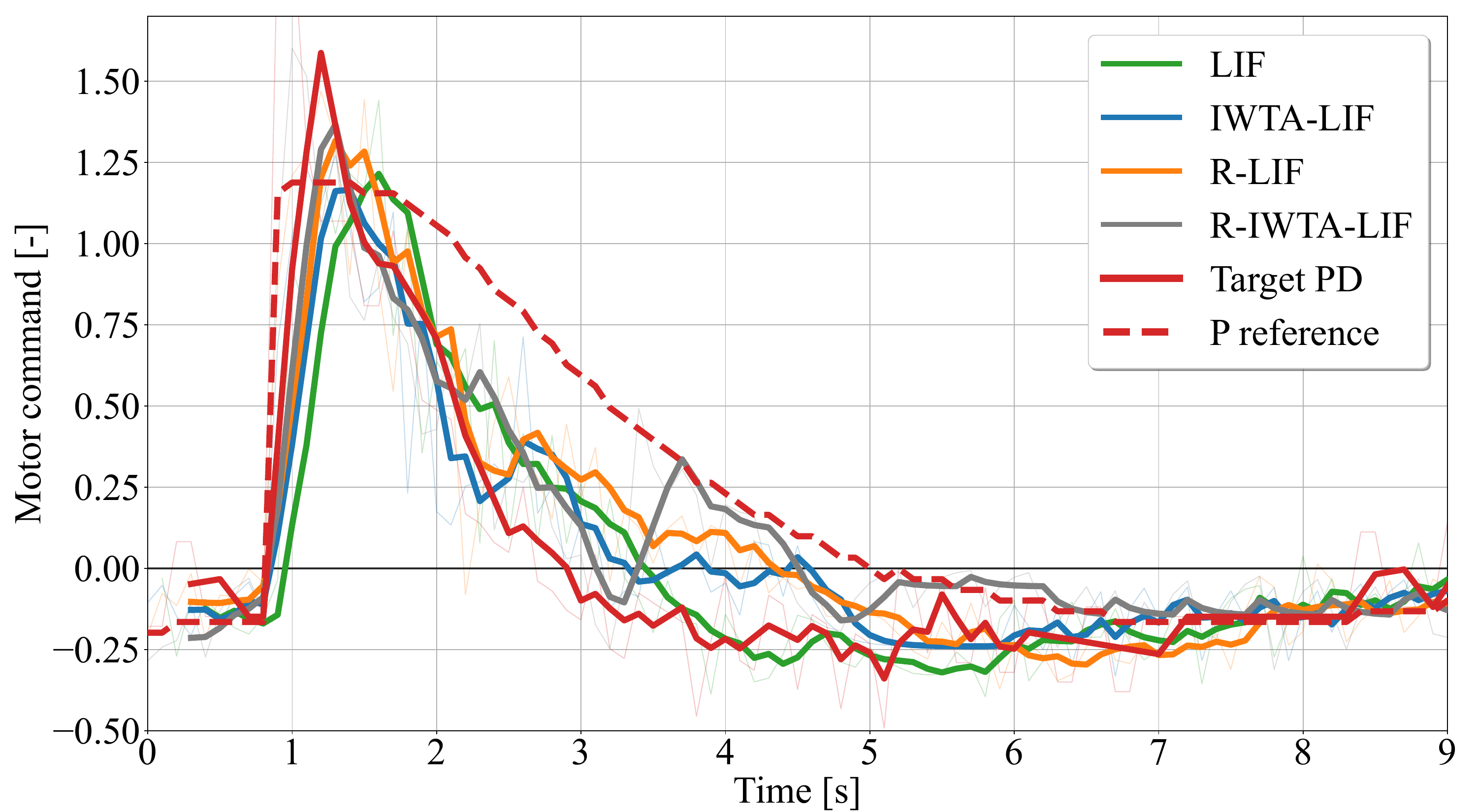}
  \caption{The moving average of the step responses of all evolved SNN controllers compared to a PD target signal using a test dataset. The LIF shows derivative action by damping the command.}
  \label{fig:PD_sim}
\end{figure}

The result of the PD training process is provided in Figure~\ref{fig:PD_sim}, showing a single step response from the test dataset.
The solid red line represents the SNN's target, while the dashed red line depicts the proportional controller. The P controller is included to visualize the additional effect of the derivative. Initially, all spiking PD controllers show clear influences of the derivative, as they match the target signal. However, after the initial damping of the proportional output, the controllers start to diverge. To prevent overshoot and oscillations, the derivative controller should counteract the proportional controller when the state is approaching the setpoint, which happens around 4 seconds in the Figure. The only controller that counteracts the P controller sufficiently is the LIF SNN. Based on this analysis and the lowest loss value across the entire test dataset, shown in Table~\ref{tab:Loss_PD_Sim}, we opted to use the LIF SNN for the blimp's altitude controller. The larger solution space for the recurrent and IWTA neuron structures makes the search space more complex and leads to local minima.

\begin{table}[h!]
\caption{Loss for SNN PD controllers ($u$) on complete test dataset (500s) using tuned PD as target ($\hat{u}$)}
\label{tab:Loss_PD_Sim}
\centering
\begin{tabular}{c|c|c|c|c|}
    & LIF   & IWTA-LIF & R-LIF & R-IWTA-LIF \\ \hline
$L(u,\hat{u})$ & 0.68 & 0.74 & 0.74   & 0.73     \\ \hline
\end{tabular}
\end{table}

\subsubsection{Integral SNN Controller}



\begin{figure}[h!]
  \centering
  \includegraphics[width = 1.0\linewidth]{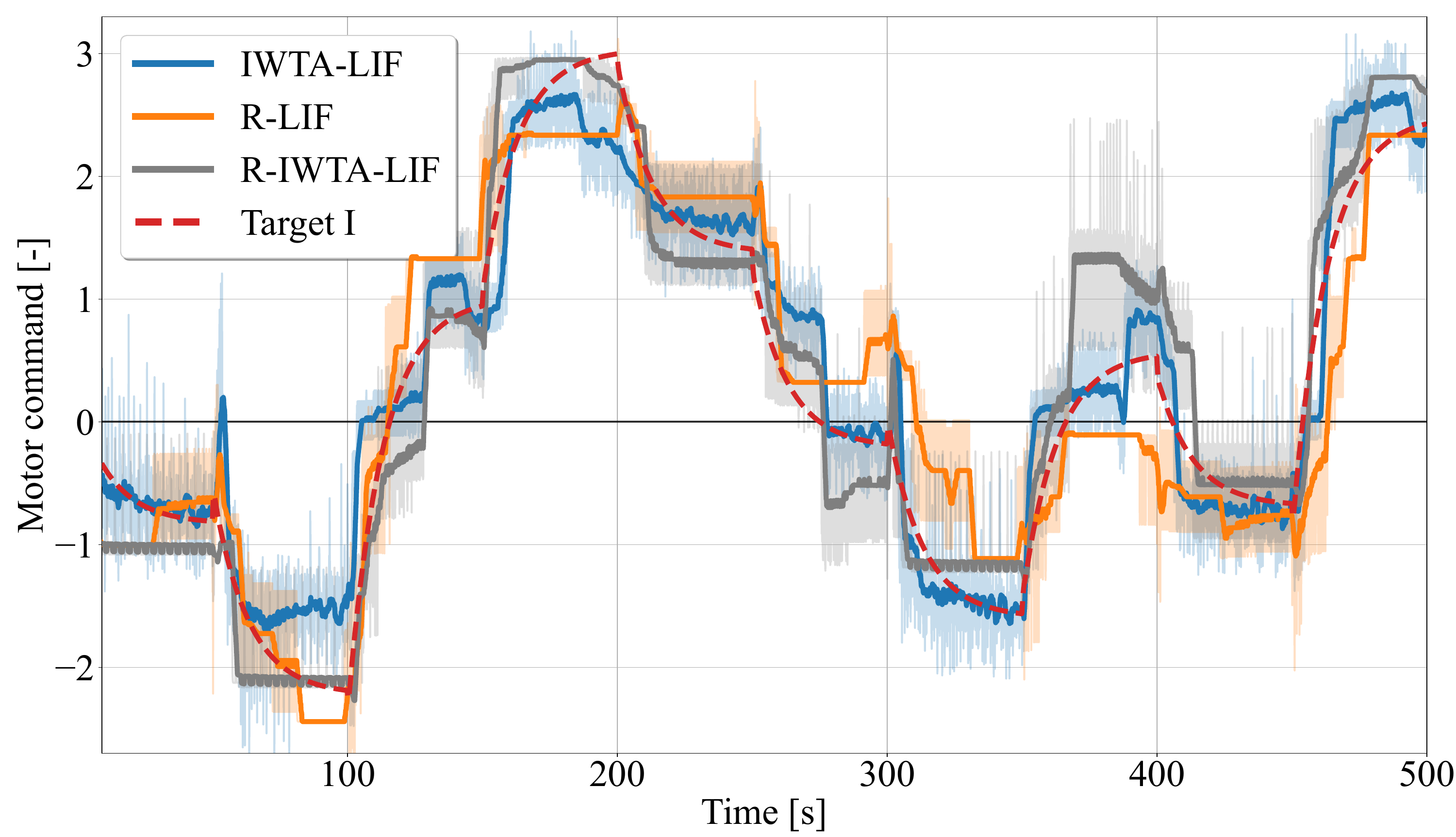}
  \caption{The moving average of the step responses of three evolved SNN controllers compared to a integral target signal using a test dataset, with a changing bias every step input.}
  \label{fig:Sim_I}
\end{figure}

The result of the integral training process is provided in Figure~\ref{fig:Sim_I}, showing the response of the different hidden layer mechanisms to a test dataset. The LIF SNN failed to learn to integrate, hence, it is excluded from the figure. 
All three spiking controllers following the test signal, without a clear standout performer. To see how well the controllers perform in the real-world, they are all tested on the indoor blimp.

\subsection{Performance of SNN controlling Real-World Blimp}
To analyze the performance of each controller separately, we first test the PD SNN controller using a neutrally buoyant blimp. Afterward, we added some weight to the blimp to make it negatively buoyant. The negatively buoyant blimp is used first to evaluate the performance of the SNN I controllers, followed by the evaluation of the fully spiking controller.

\subsubsection{PD control of neutrally buoyant blimp}
\begin{figure}[h!]
  \centering
  \includegraphics[width = 1.0\linewidth]{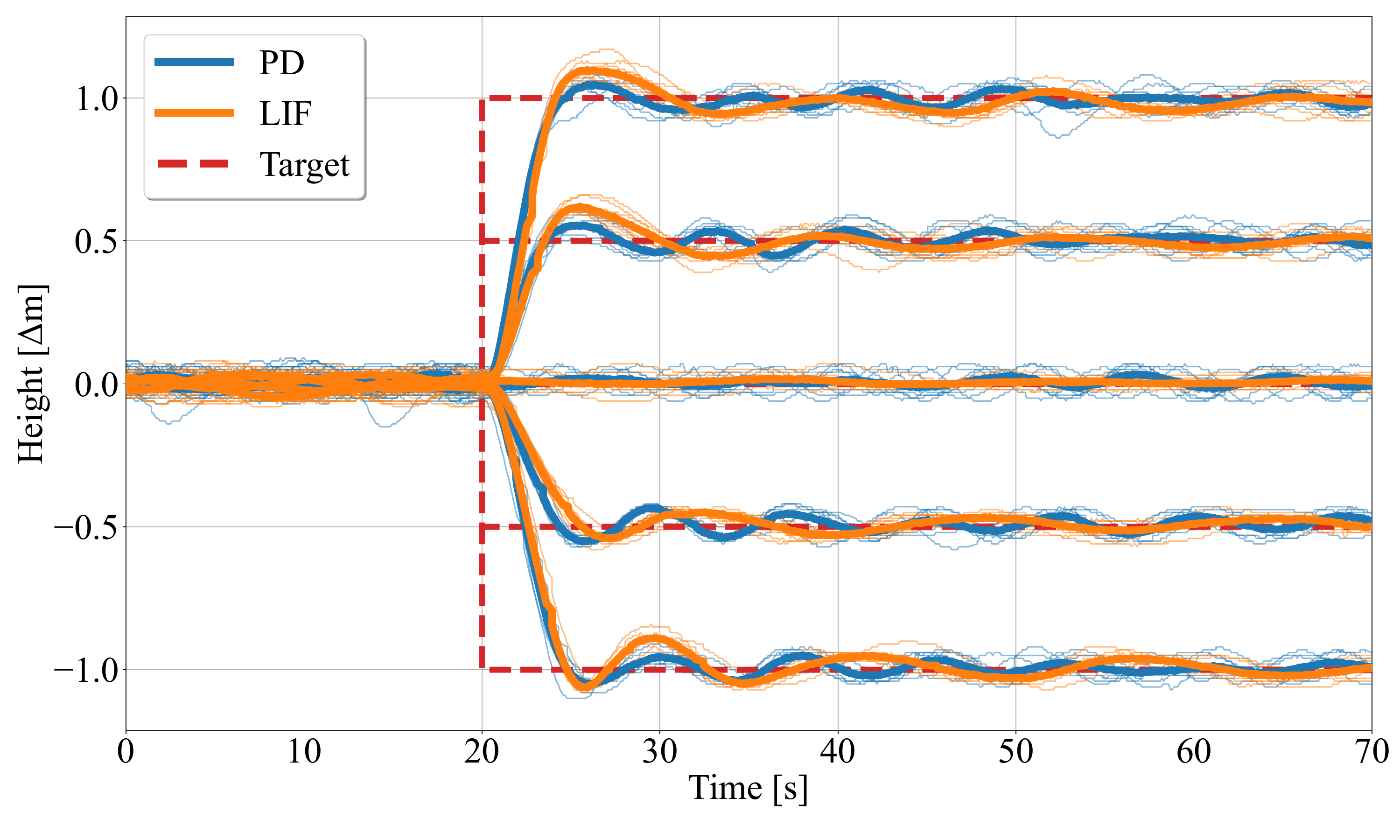}
  \caption{Real-world step responses of the altitude of the neutrally buoyant blimp using a conventional PD control and an SNN controller with no recurrency or IWTA in the hidden layer. Each setpoint was tested eight times for every controller and the average is shown by the thick line.}
  \label{fig:PD_blimp}
\end{figure}

We assess the performance of the PD SNN controller with LIF hidden structure by comparing it to a tuned conventional PD controller. The tracking accuracy is tested using five different step sizes $\Delta h$=[1,0.5,0,-0.5,-1], maintained for 50s each and the results are shown in Figure~\ref{fig:PD_blimp}. Both the conventional PD and the SNN show small oscillations, $\pm$6 cm, around the setpoint. These oscillations are caused by the discretized mapping of the motor command to the actual voltage sent to the motors using PWM. This causes a deadzone to be present in the motor command signal, which is the region of the motor commands, $u=$[-0.1,0.1], that does not result in the actuation of the rotors. 
\subsubsection{Integral control of negatively buoyant blimp}
\begin{figure}[h!]
  \centering
  \includegraphics[width = 1.0\linewidth]{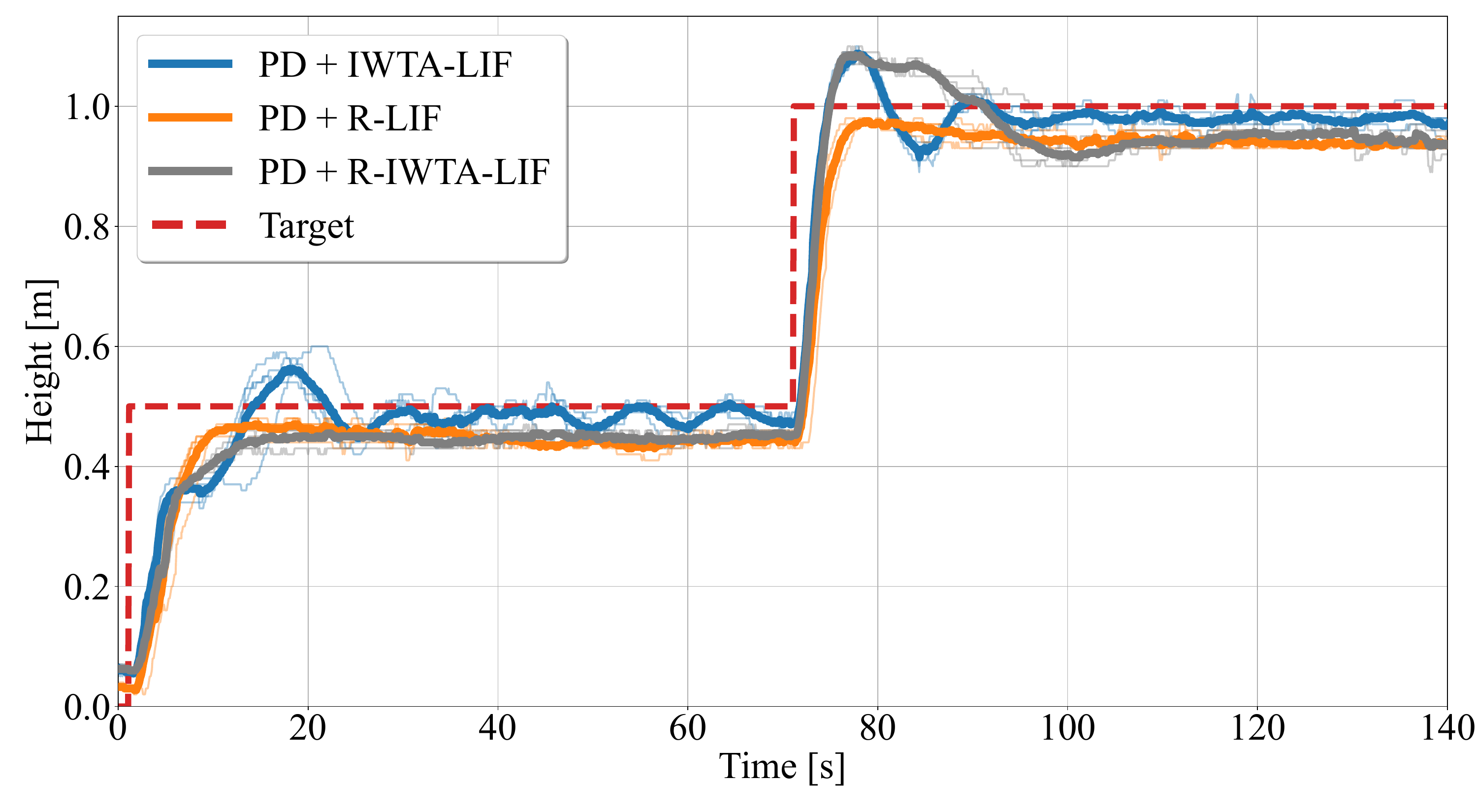}
  \caption{Real-world step responses of the altitude of the negatively buoyant blimp using a conventional PD controller and different spiking integrators. The average over five runs is shown for each controller.}
  \label{fig:I_blimp}
\end{figure}

To isolate the effect of the spiking integrator, we used the non-spiking PD controller in combination with the spiking integrator to control the negatively buoyant blimp. The result of two-step responses is shown in Figure~\ref{fig:I_blimp}, where the setpoint was changed after 70 seconds. When analyzing the steady-state error, we take the average over the last 10 seconds of each step. The steady-state error of the IWTA-LIF ($\pm$2 cm) is significantly smaller than the ones for the R-LIF and R-IWTA-LIF ($\pm$5 cm). Despite the small oscillations caused by the LIF SNN, we decided to use this spiking integrator for the full spiking controller due to the minimal steady-state error.

\subsubsection{Full spiking control of negatively buoyant blimp}
\begin{figure}[h!]
  \centering
  \includegraphics[width = 1.0\linewidth]{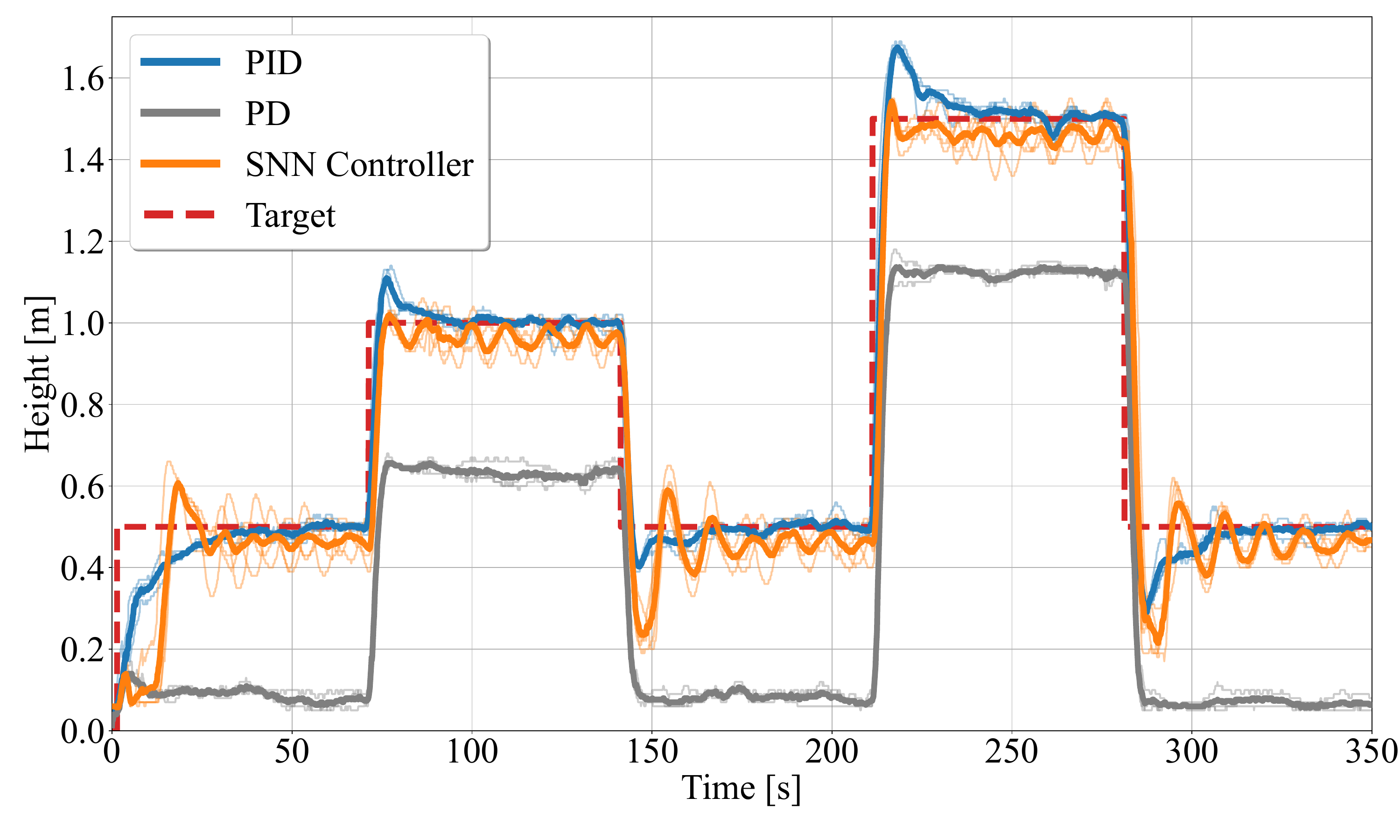}
  \caption{Real-world multi-step response of the altitude of the negatively buoyant blimp using a non-spiking PID \& PD controller and the SNN controller, which is the combination of the SNN PD (LIF) and the SNN integral (IWTA-LIF) controller. The average over five runs is shown for each controller.}
  \label{fig:Blimp_I_combined}
\end{figure}

The combination of both the spiking PD (LIF) and integral (IWTA-LIF) controller is shown in Figure~\ref{fig:Blimp_I_combined}. We analyzed the step response for the blimp using different setpoints $h$=[0.5,1,1.5]m, maintained for 70s. The combined SNN controller demonstrates effective altitude control while minimizing the steady-state error to $\pm$3cm. However, the SNN controller shows relatively large initial oscillations when receiving a downward step input, compared to the upward steps. The oscillations result from the delay introduced when the rotors must make a 180-degree turn during direction changes. Given the blimp's negative buoyancy, continuous upward thrust is essential for stability. In cases of upward step inputs, the rotors constantly push the blimp upwards. Conversely, when a lower setpoint is used, the blimp is initially pushed downwards by the rotor before pushing back up to attenuate the movement. This difference in overshoot is also visible for the PID controller, with an average overshoot of 14cm for upward steps and 20cm for downward steps.

\section{Conclusion}
In this work, we evolved a spiking neural network (SNN) that successfully controls the altitude of a non-neutrally buoyant indoor blimp. The SNN parameters were optimized through an evolutionary algorithm, facilitating extensive exploration of the solution space. This exploratory training approach allowed for an in-depth analysis of various hidden-layer configurations, recurrency and Input Weighted Threshold Adaptation (IWTA), for the Leaky-Integrate and Fire (LIF) neuron model. As a result, we developed two complementary SNN controllers which, when combined, achieved accurate tracking of the reference state. The first controller exhibited rapid response to control errors while effectively mitigating overshoot and large oscillations, after being trained on a tuned PD controller for the blimp. In parallel, the second controller was designed to minimize steady-state errors arising from the blimp's non-neutral buoyancy-induced drift. This controller learned to perform integration of the error using IWTA within the hidden layer of the network.

Despite the limitation within the blimp's current drivetrain configuration, the developed SNN controllers have showcased their ability to maintain stable altitude control, employing just 160 spiking neurons. All processing and sensing is performed onboard the blimp, with the SNN running on the Raspberry Pi's CPU. Future research will focus on the completion of the neuromorphic control loop, integrating event-based sensors with neuromorphic processors. This integration aims to fully demonstrate the potential of neuromorphic computing in robotic control.

\bibliographystyle{IEEEtran}
\bibliography{references,manual_ref}

\begin{thebibliography}{10}
\providecommand{\url}[1]{#1}
\csname url@rmstyle\endcsname
\providecommand{\newblock}{\relax}
\providecommand{\bibinfo}[2]{#2}
\providecommand\BIBentrySTDinterwordspacing{\spaceskip=0pt\relax}
\providecommand\BIBentryALTinterwordstretchfactor{4}
\providecommand\BIBentryALTinterwordspacing{\spaceskip=\fontdimen2\font plus
\BIBentryALTinterwordstretchfactor\fontdimen3\font minus
  \fontdimen4\font\relax}
\providecommand\BIBforeignlanguage[2]{{%
\expandafter\ifx\csname l@#1\endcsname\relax
\typeout{** WARNING: IEEEtran.bst: No hyphenation pattern has been}%
\typeout{** loaded for the language `#1'. Using the pattern for}%
\typeout{** the default language instead.}%
\else
\language=\csname l@#1\endcsname
\fi
#2}}

\bibitem{Zhang2022AController}
\BIBentryALTinterwordspacing
D.~Zhang, A.~Loquercio, X.~Wu, A.~Kumar, J.~Malik, and M.~W. Mueller, ``{A
  Zero-Shot Adaptive Quadcopter Controller},'' 9 2022. [Online]. Available:
  \url{http://arxiv.org/abs/2209.09232}
\BIBentrySTDinterwordspacing

\bibitem{Heryanto2017}
M.~Heryanto, H.~Suprijono, B.~Yudho, and B.~Kusumoputro, ``Attitude and
  altitude control of a quadcopter using neural network based direct inverse
  control scheme,'' \emph{Advanced Science Letters}, vol.~23, pp. 4060--4064,
  05 2017.

\bibitem{Maass1997NetworksModels}
W.~Maass, ``{Networks of spiking neurons: The third generation of neural
  network models},'' \emph{Neural Networks}, vol.~10, no.~9, pp. 1659--1671,
  1997.

\bibitem{Bing2019ANetworks}
Z.~Bing, C.~Meschede, F.~R{\"{o}}hrbein, K.~Huang, and A.~C. Knoll, ``{A survey
  of robotics control based on learning-inspired spiking neural networks},''
  \emph{Frontiers in Neurorobotics}, vol.~12, 2019.

\bibitem{Vitale2021Event-drivenChip}
A.~Vitale, A.~Renner, C.~Nauer, D.~Scaramuzza, and Y.~Sandamirskaya,
  ``Event-driven vision and control for uavs on a neuromorphic chip,'' in
  \emph{2021 IEEE International Conference on Robotics and Automation
  (ICRA)}.\hskip 1em plus 0.5em minus 0.4em\relax IEEE, 2021, pp. 103--109.

\bibitem{Taherkhani2020ANetworks}
A.~Taherkhani, A.~Belatreche, Y.~Li, G.~Cosma, L.~P. Maguire, and T.~M.
  McGinnity, ``{A review of learning in biologically plausible spiking neural
  networks},'' \emph{Neural Networks}, vol. 122, pp. 253--272, 2020.

\bibitem{Neftci2019SurrogateNetworks}
E.~O. Neftci, H.~Mostafa, and F.~Zenke, ``{Surrogate Gradient Learning in
  Spiking Neural Networks: Bringing the Power of Gradient-based optimization to
  spiking neural networks},'' \emph{IEEE Signal Processing Magazine}, vol.~36,
  no.~6, pp. 51--63, 11 2019.

\bibitem{Bing2019}
Z.~Bing, C.~Meschede, F.~R{\"{o}}hrbein, K.~Huang, and A.~C. Knoll, ``{A survey
  of robotics control based on learning-inspired spiking neural networks},''
  \emph{Frontiers in Neurorobotics}, vol.~12, 2019.

\bibitem{Stagsted2020Event-basedStudy}
R.~Stagsted, A.~Vitale, A.~Renner, and L.~B. Larsen, ``{Event-based PID
  controller fully realized in neuromorphic hardware: A one DoF study},'' in
  \emph{IEEE International Conference on Intelligent Robots and Systems},
  vol.~2, 7 2020, pp. 10\,939--10\,944.

\bibitem{Stagsted2020TowardsUAV}
R.~Stagsted, A.~Vitale, A.~Renner, and L.~Larsen, ``{Towards neuromorphic
  control: A spiking neural network based PID controller for UAV},'' in
  \emph{Robotics: Science and Systems XVI}, vol.~1, 1 2020.

\bibitem{Stroobants2022DesignProcessors}
S.~Stroobants, J.~Dupeyroux, and G.~De~Croon, ``Design and implementation of a
  parsimonious neuromorphic pid for onboard altitude control for mavs using
  neuromorphic processors,'' in \emph{Proceedings of the International
  Conference on Neuromorphic Systems 2022}, 2022, pp. 1--7.

\bibitem{Davies2018Loihi:Learning}
M.~Davies, N.~Srinivasa, T.-H. Lin, G.~Chinya, Y.~Cao, S.~H. Choday, G.~Dimou,
  P.~Joshi, N.~Imam, S.~Jain, \emph{et~al.}, ``Loihi: A neuromorphic manycore
  processor with on-chip learning,'' \emph{IEEE Micro}, vol.~38, no.~1, pp.
  82--99, 2018.

\bibitem{Zaidel2021NeuromorphicControl}
Y.~Zaidel, A.~Shalumov, A.~Volinski, L.~Supic, and E.~Ezra~Tsur,
  ``{Neuromorphic NEF-Based Inverse Kinematics and PID Control},''
  \emph{Frontiers in Neurorobotics}, vol.~15, 2 2021.

\bibitem{Stroobants2023NeuromorphicAdaptation}
S.~Stroobants, C.~De~Wagter, and G.~De~Croon, ``Neuromorphic control using
  input-weighted threshold adaptation,'' in \emph{Proceedings of the 2023
  International Conference on Neuromorphic Systems}, 2023, pp. 1--8.

\bibitem{Gonzalez-Alvarez2022EvolvedBlimp}
M.~Gonzalez-Alvarez, J.~Dupeyroux, F.~Corradi, and G.~C. De~Croon, ``{Evolved
  neuromorphic radar-based altitude controller for an autonomous open-source
  blimp},'' \emph{Proceedings - IEEE International Conference on Robotics and
  Automation}, pp. 85--90, 2022.

\bibitem{Qiu2020a}
H.~Qiu, M.~Garratt, D.~Howard, and S.~Anavatti, ``Towards crossing the reality
  gap with evolved plastic neurocontrollers,'' in \emph{Proceedings of the 2020
  Genetic and Evolutionary Computation Conference}, 2020, pp. 130--138.

\bibitem{DHPC2022_new}
{D}elft {H}igh {P}erformance {C}omputing~{C}entre ({DHPC}), \emph{{D}elft{B}lue
  {S}upercomputer ({P}hase 1)}, 2022,
  \url{https://www.tudelft.nl/dhpc/ark:/44463/DelftBluePhase1}.

\bibitem{hansen2006cma}
N.~Hansen, ``The cma evolution strategy: a comparing review,'' \emph{Towards a
  new evolutionary computation: Advances in the estimation of distribution
  algorithms}, pp. 75--102.

\bibitem{Toklu2023EvoTorch:Python}
\BIBentryALTinterwordspacing
N.~E. Toklu, T.~Atkinson, V.~Micka, P.~Liskowski, and R.~K. Srivastava,
  ``{EvoTorch: Scalable Evolutionary Computation in Python},'' pp. 1--25, 2023.
  [Online]. Available: \url{http://arxiv.org/abs/2302.12600}
\BIBentrySTDinterwordspacing

\bibitem{Benesty2009}
I.~Cohen, Y.~Huang, J.~Chen, J.~Benesty, J.~Benesty, J.~Chen, Y.~Huang, and
  I.~Cohen, ``Pearson correlation coefficient,'' \emph{Noise reduction in
  speech processing}, pp. 1--4, 2009.

\end{thebibliography}

\end{document}